\documentclass[conference]{IEEEtran}
\IEEEoverridecommandlockouts
\usepackage{tikz}
\usepackage{cite}
\usepackage{amsmath,amssymb,amsfonts}
\usepackage{algorithmic}
\usepackage{graphicx}
\usepackage{textcomp}
\usepackage{xcolor}

\usepackage{booktabs}
\usepackage{booktabs}
\def\BibTeX{{\rm B\kern-.05em{\sc i\kern-.025em b}\kern-.08em
    T\kern-.1667em\lower.7ex\hbox{E}\kern-.125emX}}
\begin{document}

\title{A contrastive-learning approach for auditory attention detection
\thanks{This work was supported in part by the National Science Foundation under grant IIS-2235228, the Center for Cognitive and Brain Sciences, and by the Ohio Supercomputer Center.}
}

\author{\IEEEauthorblockN{ Seyed Ali Alavi Bajestan}
\IEEEauthorblockA{\textit{Computer Science and Engineering} \\
\textit{The Ohio State University}\\
Columbus, OH, USA \\
alavibajestan.1@osu.edu}
\and
\IEEEauthorblockN{Mark Pitt}
\IEEEauthorblockA{\textit{Psychology} \\
\textit{The Ohio State University}\\
Columbus, OH, USA \\
pitt.2@osu.edu}
\and
\IEEEauthorblockN{Donald S. Williamson}
\IEEEauthorblockA{\textit{Computer Science and Engineering} \\
\textit{The Ohio State University}\\
Columbus, OH, USA \\
williamson.413@osu.edu}

}
\maketitle
\begin{abstract}
Carrying conversations in multi-sound environments is one of the more challenging tasks, since the sounds overlap across time and frequency making it difficult to understand a single sound source. One proposed approach to help isolate an attended speech source is through decoding the electroencephalogram (EEG) and identifying the attended audio source using statistical or machine learning techniques. However, the limited amount of data in comparison to other machine learning problems and the distributional shift between different EEG recordings emphasizes the need for a self supervised approach that works with limited data to achieve a more robust solution. In this paper, we propose a method based on self supervised learning to minimize the difference between the latent representations of an attended speech signal and the corresponding EEG signal. This network is further finetuned for the auditory attention classification task. We compare our results with previously published methods and achieve state-of-the-art performance on the validation set.
\end{abstract}
\section{Introduction}
\label{sec:intro}

Everyone  struggles with understanding speech in noisy environments, such as at  parties or sporting events.  
Many machine learning approaches, based on speech enhancement or speaker extraction, have been proposed to isolate the attended sound signal amongst competing noise or talkers, respectively. These approaches, however, assume that the attended signal is known beforehand, which is not the case in reality. 

Auditory attention detection (AAD) addresses this problem, where statistical and machine learning approaches have been developed. The general idea is that the electroencephalographic (EEG) signal can be provided as input to a machine learning or statistical algorithm that uses the input to determine the attended sound source of the listener. The pivotal work in \cite{cherry1953some} introduced the idea of the dependency of brain processing in cocktail party scenarios on many factors including, the location of the voice, visual queues like lips-reading, and the speaker's pitch, to name a few.  In \cite{mesgarani2012selective}, it is shown that the spectral reconstruction of the cortical surface response of multiple speakers were highly correlated with the temporal and spectral features of the attended signal, where low correlation results for the unattended signal. Leveraging this finding, many early AAD approaches reconstruct the audio stimuli from the EEG signal (backward or reverse approaches) and identify the audio source with the highest correlation to the EEG signal as the attended source \cite{rieke1995naturalistic, stanley1999reconstruction, mesgarani2009influence, pasley2012reconstructing, golumbic2013mechanisms}.
These approaches are known as linear methods, which also include correlation based learning, canonical correlation analysis (CCA) \cite{de2018decoding}, and linear filter based methods (e.g. FIR) \cite{alickovic2019tutorial}. The simplicity and low computational cost of these models are, however,  accompanied with low generalizability and accuracy, especially for small decision windows, which is undesirable. 

More recently, approaches based on nonlinear mapping \cite{de2020machine} and convolutional neural networks (CNNs) \cite{vandecappelle2021eeg, cai2020low, ciccarelli2019comparison} have been developed for auditory attention detection. A comparison of linear based methods versus neural network based methods \cite{ciccarelli2019comparison} shows a drastic improvement (about 15\% increase in average accuracy) in detection for neural network based methods, even with a lower number of recording electrodes. 
%
%
A CNN based approach for similarity maximization of the latent space of the EEG and sound sources was used in \cite{geirnaert2021electroencephalography}, which tried to maximize the similarity with the attended speech. A recursive gated convolution method for left and right classification using only the EEG data was used in \cite{cai2023rgcnet}, where graph convolutional networks have also been used \cite{wangeeg}. In \cite{cai2021eeg}, frequency and channel based attention is used for the EEG data, where a CNN network helps to predict AAD. Similarly, the work in \cite{kuruvila2021extracting} used a joint LSTM-CNN model. Many alternative neural network approaches have been proposed including, graph neural networks \cite{cai2023brain}, which have the advantage of better signal explainability, but at the cost of accuracy; cross modal attention on the EEG and audio frequency bands \cite{li2021biologically}; attention \cite{su2022stanet}; and an unsupervised learning approach that showed a label-independent learning stage can be used to surpass a supervised only approach  \cite{geirnaert2021unsupervised}.   
%
%
To the best of our knowledge, however, only one approach evaluates AAD performance on an unseen subject whose data is excluded from the training set \cite{beauchene2023subject}, which is the optimal way to assess how well the approach generalizes.  

\begin{figure*}[htb!]
    \centering
    \includegraphics[width=0.8\linewidth]{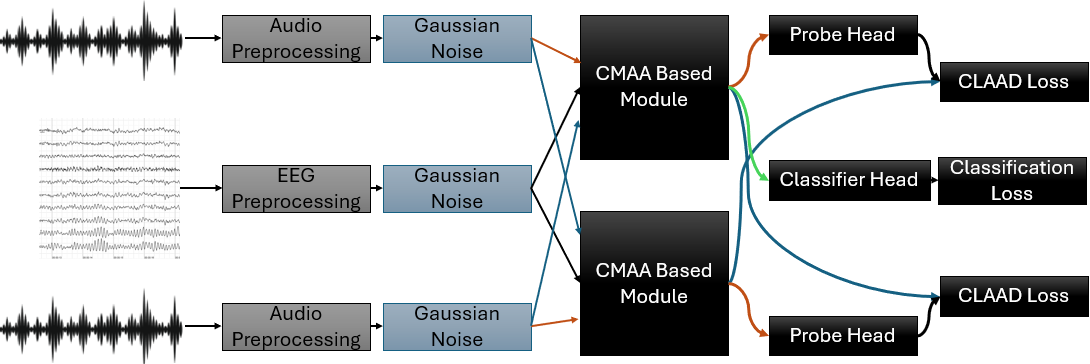}
    \caption{A high-level depiction of the proposed network. The preprocessed audio and EEG data have a Gaussian noise added to them. Each triplet is fed to the two CMAA encoders with parameter sharing. The encoder outputs are provided to two sets of probe and classification heads. The probe head receives a representation from the CMAA encoder, where a shallow network was chosen so the CMAA module would be as descriptive as possible. The classification heads each find a boundary within the latent representation space. 
    Two CLAAD losses and a classification loss are then computed.
    }
    \label{fig:1}
\end{figure*}

 Although convolutional and conventional neural network approaches have advantages over linear methods, these approaches are 
 time dependant or they do not capture the essential differences between electrodes. Contrastive learning, as one of the many different methods for self supervised learning, learns the internal patterns and structures in the data, without a training label. Inspired by this approach, we propose a new method to learn the EEG patterns and make the learned representation more useful for the downstream task of attended source detection. 
As the base of our method, we use a pair of cross-modal attention-based auditory attention detection (CMAA) encoders, introduced in \cite{cai2021auditory}, which uses a cross modal attention module \cite{vaswani2017attention}. 
We devise a loss function that maximizes the similarity of the encoder representations out of the two CMAA paths, and minimize the binary cross entropy divergence for the classification loss. This approach addresses the need for better generalization in between experiments and between subject, one of the challenges in the biological signal processing. A contrastive approach has been proposed \cite{chen2023auditory}, but it uses a variational auto encoder (VAE) for latent representation learning, which even though it encourages more flexibility over an auto encoder, it is not a better choice over CMAA for a contrastive approach, due to the added stochasticity.

\section{Proposed Approach}

A depiction of our proposed approach is shown in figure \ref{fig:1}. Common spatial pattern (CSP) features are first computed from the EEG signals \cite{koles1990spatial}, after pre-processing the EEG data (more in section \ref{prep}). The EEG recordings have a high correlation with nearby electrodes, which is an issue because it increases redundancy in the data and limits the useful information available to the neural network. CSP address this by maximizing the variance between different categories which in turn makes more distinctive inter-channel EEG data available. CSP has been used for feature processing for several approaches \cite{cai2020low, yang2023auditory}, and it has helped improve performance. More details can be found in \cite{koles1990spatial}. 

The audio envelope is extracted after bandpass filtering. The attended and competing audio sources are both processed in this manner. The processed audio signals are each paired with the procesesed EEG data, and provided to separate CMAA encoders, with the goal of getting meaningful contrastive and classification projections from their latent representations that can be useful for the AAD classification task. By utilizing self attention modules in the CMAA modules, we get better generalization in comparison with CNN blocks alone.

The network is based on the contrastive method. In \cite{chen2020simple}, a contrastive method was proposed in which the latent representation produced from two augmentations of the same data instance should be close together, while representations for augmentations of different data instances should be farther apart. This method requires large batch sizes and a sizable dataset, which is not applicable to this work, mainly due to the limited size of available AAD datasets. In order to overcome this limitation, a new self supervised approach was adapted \cite{chen2021exploring}, where the base network was the same, but it uses different augmentations of the same input that are passed through different paths of the network that share parameters. In our experiments, it was shown that the augmentation was essential for the training.

The data augmentation for the EEG is challenging. Unlike the data augmentation applied to images, where we have multiple transformation options, which after they are applied, can generate valid and in distribution data examples, for EEG data we do not have an intuitive understanding of appropriate transformations that produce valid EEG signals. To mitigate this issue, we used standard normal Gaussian noise (e.g., zero mean and unit variance) as an augmentation for the audio and the CSP of the EEG data, across all channels and time samples, to improve generalization. Gaussian noise is chosen since it does not change the expectation of the signal, but increases its variance. 

 \begin{figure}[ht]
    \centering
        \centering
        \includegraphics[width=0.48\linewidth]{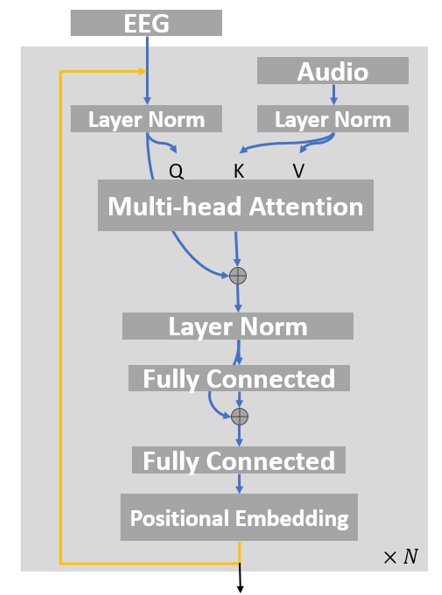}
    \caption{The cross attention module takes the preprocessed EEG and audio at the input at the first iteration. In the subsequent iterations the EEG input is replaced by the output of the previous stage. }
    \label{fig:multinetwork}
\end{figure}
tion{CMAA Encoder}
The CSP feataugmented ures are thenand audio  provided as inputs to our network. The base network is composed of five components shown in figure \ref{fig:multinetwork}. For each CMAA sub component, we layer normalize the audio and EEG input, and provide the normalized audio and EEG outputs to a multi-head attention (MHA) block within the cross attention module. The MHA output is added with the query input using a skip connection. Next the output is passed through a normalization layer, a fully connected (FC) layer, and then the FC output is added with the FC input using a skip connection. Another FC layer follows. Next to add the positional information, the final FC output is added to the positional encoding  \cite{vaswani2017attention} to keep the network informed about the sequential ordering. This creates the output for the first EEG-audio pairing, but this process is repeated $N=5$ times here and for the other EEG-audio pairing. Also, eight attention heads were used. This is performed for each audio signal and EEG data pairing, where the two outputs are concatenated together, and then provided to a fully connected layer. We experimented with using cosine similarity between the different audio and EEG pairs instead of performing concatenation, but concatenation produced better results. The representations are then provided to pairs of probe and classifier heads, which are each three FC layers.

\subsection{Contrastive Learning}

The output of one CMAA encoder is projected to a function of the second CMAA encoder which received inputs with different noise values. In \cite{chen2020simple}, the output is treated as a class category in a batch, so in a batch of size B, a cross entropy was calculated for a class size B, meaning that each latent representation should maximally match it's augmented version after passing through the network. Here we took a different route, in order to mitigate the limitation imposed by a small batch size and also minimize the effect of variability between different recordings. More specifically, we define a loss function that tries to minimize the distance between different recordings with the same level of added noise. This way, not only the distance between the data and its augmentation is minimized, its distance to the augmentation of different recordings of the same label is also reduced. The cost functions are below
\begin{align}
    &\mathcal{L}_{\text{Classification Loss}} = -\sum_{i=1}^{\mathcal{A}} \sum_{c=1}^{2} y_{i,c} \log(\hat{y}_{i,c})\hfill\\
    &\mathcal{L}_{\text{positive pair}}(p,z,y) =  \dfrac{\Sigma_{i \in \mathcal{A}} e^{p\times z^T} . (y_{one hot}\times y_{one hot}^T)}{\Sigma_{j \in \mathcal{A}} e^{p\times z^T}}\\
    &\mathcal{L}_{\text{CLAAD}}= \dfrac{1}{2}(\mathcal{L}_{\text{positive pair}}(p_1,z_2,y) + \mathcal{L}_{\text{positive pair}}(p_2,z_1,y))
    \label{eq:CLAAD}
\end{align}
where in equation \eqref{eq:CLAAD}, which we will call the Contrastive Learning Auditory Attention Detection (CLAAD) loss, $p$ refers to the output of the encoder head of one path and $z$ refers to the output of the other encoder path that has passed through a contrastive probe head. $\mathcal{A}$ refers to the samples in a batch and $y$ refers to the label. $\hat{y}$ refers to the predicted probability and $y$. $y_{onehot}$ also refers to the one-hot vector of the class labels. $p_1$ and $p_2$ refers to the output of the Probe Head with two different noise samples at the input of the CMAA block. The same goes for the $z_1$ and $z_2$ which refer to the output of the CMAA block.

The contrastive loss (e.g., CLAAD) and the classifier loss are trained back-to-back. Meaning the loss of the contrastive part is calculated, it's gradient is applied through the optimizer then the process is repeated for the classifier loss. In contrast to the common method of freezing the encoder weights during the classifier training, we allowed them to be updated. We observe that this approach results in higher accuracy in comparison with frozen weights. Furthermore, the classifier is trained using the cross-entropy loss.

\section{Experiments}

\subsection{Data}

We train and evaluate our approach using the DTU dataset \cite{wong2018comparison}, which is commonly used for AAD. The data consists of EEG recordings of 18 subjects while listening to pairs of audio stimuli. The stimuli contains two speakers (male and female) reading a story in an anechoic chamber,  
which were lateralized in the $+60$ and $-60$ along the azimuth direction at $2.4$ meters. The audio was presented in three acoustic conditions, non echoic, mildly reverberant, and high reverberation. These conditions were pseudo randomized during the experiment. In each experiment, the subjects listened to the recordings and answered questions about the content. The data was recorded using a Biosemi Active 2 system, with 64 channels and at 512 sampling rate. Six additional electrodes were used at mastoids, and for vertical and horizontal electrooculogram. 


\subsection{Data Preprocesssing}
\label{prep}

We apply power law compression to the audio data \cite{biesmans2016auditory}, since it can significantly improve AAD performance. In this approach, gammatone filters between 150 to 4000 Hz filter audio into subbands, where each subband is then raised to the power of 0.6. The subbands are recombined to generate the broadband signal. Then it was filtered further before being downsampled to 64 Hz to 
\begin{table}[th]
  \caption{Per-subject accuracy with 5-fold cross validation.``-" denotes unreported results in the original papers. }
  \label{tab:ex}
  \centering
  \begin{tabular}{| r@{} ||c |c|c|c|c |c|c|}
    \toprule
    \multicolumn{1}{c}{\textbf{Model}} &  
    \multicolumn{1}{c}{\textbf{0.1s}} & 
    \multicolumn{1}{c}{\textbf{0.5s}}  & 
    \multicolumn{1}{c}{\textbf{1s}}  &\multicolumn{1}{c}{\textbf{2s}}  & 
    \multicolumn{1}{c}{\textbf{3s}}  & \multicolumn{1}{c}{\textbf{5s}}  & \multicolumn{1}{c}{\textbf{30s}}\\
    \hline
     CCA \cite{geirnaert2021electroencephalography}  &-&-&60& - & - & 70 &84\\
      RGCnet \cite{cai2023rgcnet} & 66.4 & 72.1 &76.9 & 87.6 & - &-&-\\
     ST-GCN \cite{wangeeg} & 68.5&75.4 &77.3 & - & -&-&-\\
    STAnet \cite{su2022stanet} &66 & 71 &72& 74&- & 77 & -\\
    \cite{wong2018comparison}  & - & - & 52 & -&62&-&82\\ 
    CNN-CF \cite{cai2021eeg} &-& - & 79.3 & 82.9 & -&-&-\\
    \cite{geirnaert2021unsupervised} & - & - & 62&-&-&68 & -\\
    \cite{kuruvila2021extracting} & -&-&- & 55 & 55 & 55& -\\
     EEG-Graph \cite{cai2023brain}& - & - & 63.3 &- &-&-&-\\
     BIAnet \cite{li2021biologically}& 75.2&78.1 &79.0& 80.6 & - & -&-\\
     TMC-VAE \cite{chen2023auditory}& - & - & - & 80.8&92.1&-&-\\
    \textbf{CLAAD (ours)}   & - & \textbf{85.96} & - & \textbf{96.72} & - & \textbf{86.76} & -\\
\hline
    \bottomrule
  \end{tabular}
\end{table}
match the downsampled EEG signal. The CSP transform has 64 dimensions.
%
%
The EEG data was first re-referenced by the Cz electrode (positioned over the corpus callosum). Then it was bandpassed filter between 1 and 9 Hz, before being downsampled to 64 Hz.

\subsection{Network Parameters}
The contrastive step and the classification steps were trained using the Adam optimizer \cite{kingma2014adam} with  a learning rate of 0.001, $\beta_1$ of 0.9, and $\beta_2$ 0f 0.99. The model was trained for 40 epochs. The CMAA output dimension is 50, the same as the probe output dimension, the first FC layer has a RELU activation while the second one has linear activation. The middle layer of the probe head is 25. The fully connected layers of the classifier have 
dimensions of 100, 50, and 2, respectively. The dimension of the fully connected layers in the cross attention module is 320, the same goes for all the fully connected layers in figure \ref{fig:multinetwork}. Eight attention head are used.

\section{Results}

 The data was divided into decision windows of 0.5, 2 and 5 second lengths with $50\%$ overlap, before being fed to the network for training, which are commonly-used window lengths. 
We ran two experiments. Eighty percent of the data was used for training and $20\%$ for validation. First we use 5-fold cross validation to train and validate the model for each subject and reported the average accuracy among them. Second, we use cross-subject validation, in which the network was trained on all subjects excluding one subject, and the trained network was then tested on the excluded subject.

\begin{figure}
    \centering
    \includegraphics[width=0.85\linewidth]{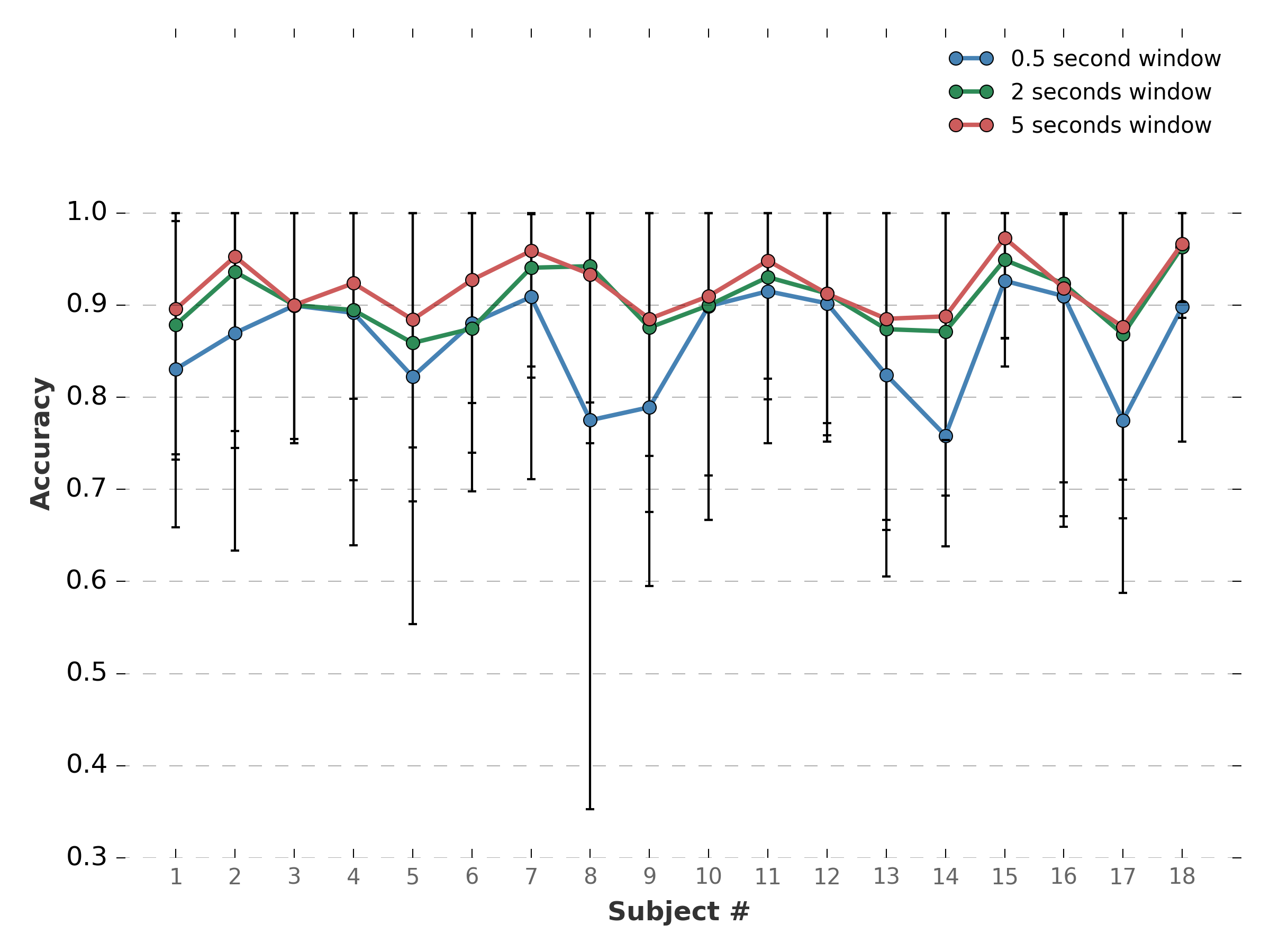}
    \includegraphics[width=0.85\linewidth]{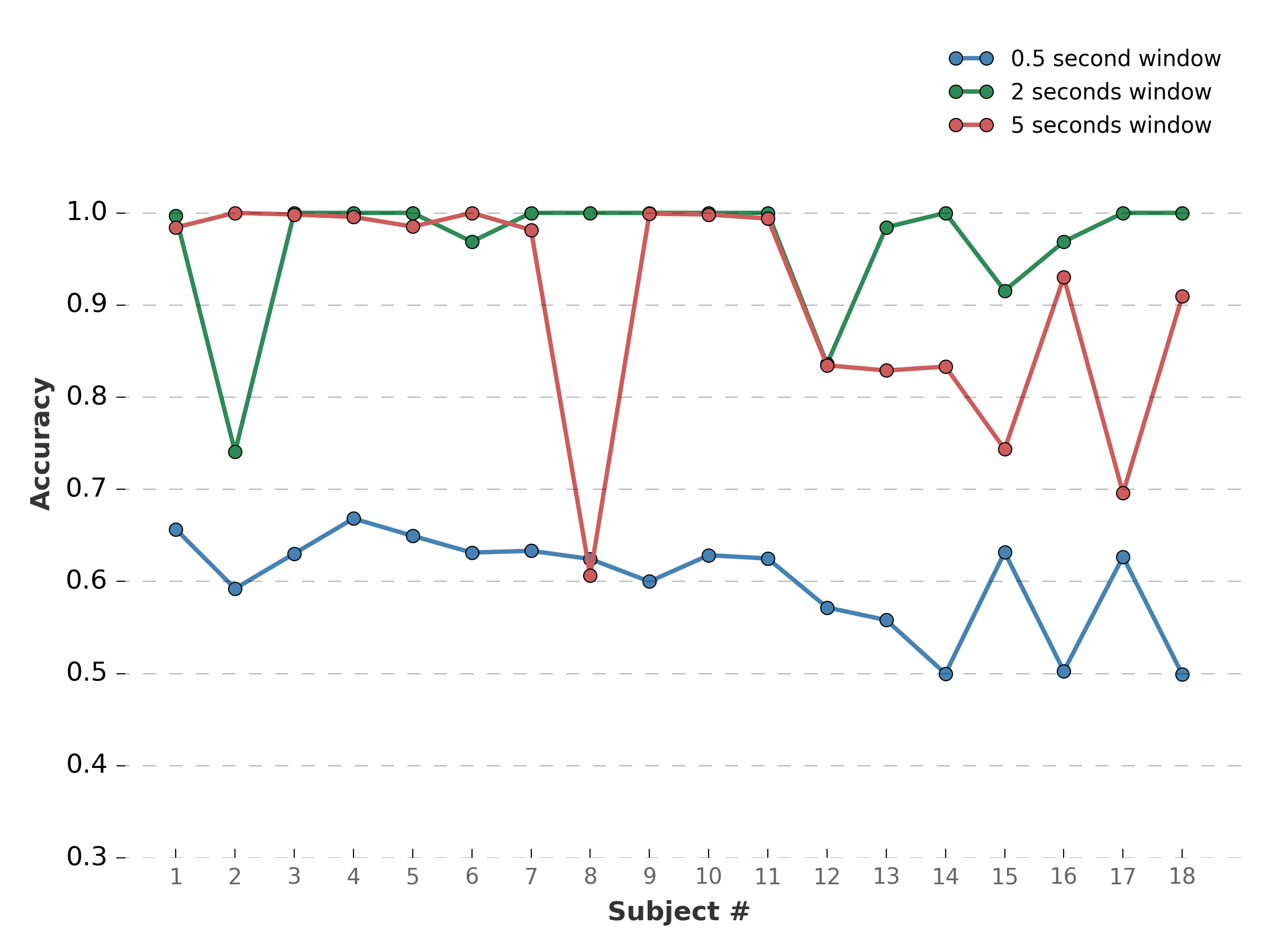}
    \caption{The average per 
subject accuracy for the validation set (Top),  where the error bars represent the maximum and minimum accuracy of different folds. The results when the subject data is unseen is shown below (bottom).}
    \label{fig:persubject}
\end{figure}

Table \ref{tab:ex} shows the per-subject accuracy of the network trained on the DTU dataset. We achieved the top accuracy on the 0.5s, 2s and 5 second windows among different methods. The comparison results are as reported in the corresponding papers. In \cite{beauchene2023subject}, the authors report 63\% accuracy for the 5 second window. We achieved better results for this window size and also achieved comparably high accuracy for the 2 seconds window. 
This generalization is attributed to the CLAAD loss, which helps the network learn data patterns rather than finding a close model for the classification task alone. The results for each subject using our proposed approach are shown in figure \ref{fig:persubject} (top), where performance is relatively consistent for each subject for all decision windows.

In figure \ref{fig:persubject} (bottom) the result for the exclusion case is plotted, in which one subject is reserved for validation and the data of the other subjects is used for training. 
%
%
%
The figure shows that in some cases the model gets almost perfect validation accuracy, especially for the 2 and 5 second decision windows. The 0.5 decision produces the lowest results, which is expected since it is the most challenging scenario. Note that these results are still better than the 5 second decision window results from another approach that held data from one subject for evaluation (e.g. \cite{beauchene2023subject} with $63\%$ accuracy). 

\section{Conclusion}
In this paper we proposed a self supervised approach based on contrastive learning to produce a more meaningful representation of the input data which can help in a downstream AAD task. To show that a self supervised approach performs better than conventional neural network with cross entropy loss, we adopted the CMAA network as the base encoder and showed with minimal change of parameters, we can have a much more effective network. To show the generalizability of our method, we tested the network on unseen training subjects, which shows even in presence of distribution shift in between subject, the network learns good generalizations in most cases.


\bibliographystyle{ieeetr}
\bibliography{refs}

\end{document}